\documentclass[conference]{IEEEtran}
\IEEEoverridecommandlockouts
\usepackage{cite}
\usepackage{amsmath,amssymb,amsfonts}
\usepackage{setspace}
\usepackage{algorithm}
\usepackage{algorithmic}
\usepackage{graphicx}
\usepackage{textcomp}
\usepackage{xcolor}
\usepackage{multirow,booktabs,threeparttable}
\def\BibTeX{{\rm B\kern-.05em{\sc i\kern-.025em b}\kern-.08em
    T\kern-.1667em\lower.7ex\hbox{E}\kern-.125emX}}
\begin{document}

\title{Communication-Efficient Consensus Mechanism for Federated Reinforcement Learning}

\author{\IEEEauthorblockN{Xing Xu}
\IEEEauthorblockA{\textit{Zhejiang University} \\
hsuxing@zju.edu.cn}
\and
\IEEEauthorblockN{Rongpeng Li$\ast$}
\IEEEauthorblockA{\textit{Zhejiang University} \\
lirongpeng@zju.edu.cn}
\and
\IEEEauthorblockN{Zhifeng Zhao}
\IEEEauthorblockA{\textit{Zhejiang Lab} \\
zhaozf@zhejianglab.com}
\and
\IEEEauthorblockN{Honggang Zhang}
\IEEEauthorblockA{\textit{Zhejiang University} \\
honggangzhang@zju.edu.cn}
}

\maketitle

\begin{abstract}
The paper considers independent reinforcement learning (IRL) for multi-agent decision-making process in the paradigm of federated learning (FL). We show that FL can clearly improve the policy performance of IRL in terms of training efficiency and stability. However, since the policy parameters are trained locally and aggregated iteratively through a central server in FL, frequent information exchange incurs a large amount of communication overheads. To reach a good balance between improving the model's convergence performance and reducing the required communication and computation overheads, this paper proposes a system utility function and develops a consensus-based optimization scheme on top of the periodic averaging method, which introduces the consensus algorithm into FL for the exchange of a model's local gradients. This paper also provides novel convergence guarantees for the developed method, and demonstrates its superior effectiveness and efficiency in improving the system utility value through theoretical analyses and numerical simulation results.
\end{abstract}

\begin{IEEEkeywords}
Independent Reinforcement Learning, Federated Learning, Consensus Algorithm, Communication Overheads, Utility Function
\end{IEEEkeywords}

\section{Introduction}
With the development of wireless communication and advanced machine learning technologies in the past few years, a large amount of data has been generated by smart devices and can enable a variety of multi-agent systems, such as smart road traffic control \cite{Ata2020Adaptive}, smart home energy management \cite{A2017A}, and the deployment of unmanned aerial vehicles (UAVs) \cite{Schwarzrock2018Solving}. Through deep reinforcement learning (DRL), an intelligent agent can gradually improve the performance of its parameterized policy via the trial-and-error interaction with the environment. Applying DRL directly to multi-agent systems commonly faces several challenging problems, such as the non-stationary learning environment and the difficulty of reward assignment \cite{Arulkumaran2017Deep}. As an alternative, independent reinforcement learning (IRL) is often employed in practical applications to alleviate the above-mentioned problems, where each agent undergoes an independent learning process with only self-related sensations \cite{Tan1993Multi}. 

For each IRL agent, the training samples are obtained by going through a trajectory with the predefined terminal state or a certain number of Markov state transitions from Markov decision processes (MDPs). With the obtained samples, an agent can calculate the policy gradients and improve its performance by updating its parameters along the gradient descent direction. Typically, performance of the trained policy is closely related to the amount and variety of obtained samples, since the more fully explored state space leads to more accurate estimation of the cumulative reward signal. On the other hand, federated learning (FL) is a parallelly distributed machine learning paradigm, aiming to train specific model through the samples distributed across different devices. In this paper, we model FL in IRL by indicating each device with an agent, and believe that FL can help to indirectly enrich the sample information of each IRL agent through the included gradients aggregation phase, thus improving the policy performance of agents.

Since agents in IRL are naturally distributed, the policy gradients are calculated locally and need to be shared among the multi-agent system through a coordination channel. In the paper, this coordination channel is achieved by deploying a central server, aiming to iteratively aggregate the policy gradients from agents and provide new parameters to update their policies. However, there may be a large number of agents or policy iterations during the training phase. This naive implementation of FL may cause excessive communication overheads between agents and the central server. Recent studies \cite{Xing2019Stigmergic,Cha2020Proxy} which try to combine distributed DRL with FL mostly focus on improving the involved agents' capability and their collaboration efficiency, while these excessive communication overheads are ignored. In this paper, we introduce the periodic averaging method in FL to alleviate this problem \cite{Hao2018Parallel}, in which agents are allowed to perform several local updates to the model within a period before their local gradients are transmitted to the central server for averaging. The periodic averaging method can reduce the frequency of information exchange between agents and the central server, thus decreasing the corresponding communication overheads dramatically. However, an increase in the number of local updates becomes essential to guarantee the model's convergence performance. Therefore, appropriate optimization methods should consider a better balance between reducing communication overheads and improving convergence performance.

To jointly take account of the model's error convergence bound and the mainly required communication and computation overheads during the training phase, this paper proposes a system utility function and develops a consensus-based optimization scheme on top of the periodic averaging method. We introduce the consensus algorithm \cite{R2007Consensus} into both IRL and FL, where agents are allowed to exchange their models' local gradients with neighbors directly before performing each local update. The consensus algorithm has been applied in several decentralized averaging methods \cite{Jiang2017Collaborative,Xiangru2018Asynchronous}, which normally disable the central server's functionality but allow agents to average their models' parameters directly with neighbors. These decentralized averaging methods will highlight their advantages in terms of the communication overheads when the cost of devices-to-devices is much less than that of devices-to-server. However, decentralized averaging methods are unable to take full advantage of FL, and will slow the convergence speed as the size of the network of agents increases. Therefore, in the consensus-based method, we combine FL with the consensus algorithm to guarantee the convergence performance. We also theoretically show that the consensus-based scheme can reduce the model's error convergence bound dramatically. Different from the spectral radius of a matrix applied in theoretical analyses about the consensus algorithm in FL \cite{Multi-Stage2020Seyyedali}, this paper gives the model's error convergence bound under the consensus-based method from the perspective of the algebraic connectivity of the graph comprised by agents and their connections, which can additionally reflect the effect of interaction metric (i.e., local interaction step size $\epsilon$) in the consensus algorithm on convergence results. Furthermore, through a multi-agent reinforcement learning (MARL) simulation scenario, we demonstrate the superiority of the developed method in terms of improving the convergence performance and system utility value.  

The remainder of this paper is organized as follows. In Section \uppercase\expandafter{\romannumeral2}, we introduce the system model, formulate the optimization problem, and define a system utility function. In Section \uppercase\expandafter{\romannumeral3}, we describe the developed optimization method and analyze its error convergence bound. In Section \uppercase\expandafter{\romannumeral4}, we introduce a MARL simulation scenario, and present the corresponding results of the developed method. In Section \uppercase\expandafter{\romannumeral5}, we conclude this paper with a summary. 

\begin{figure}
	\centering
	\includegraphics[width=0.49\textwidth]{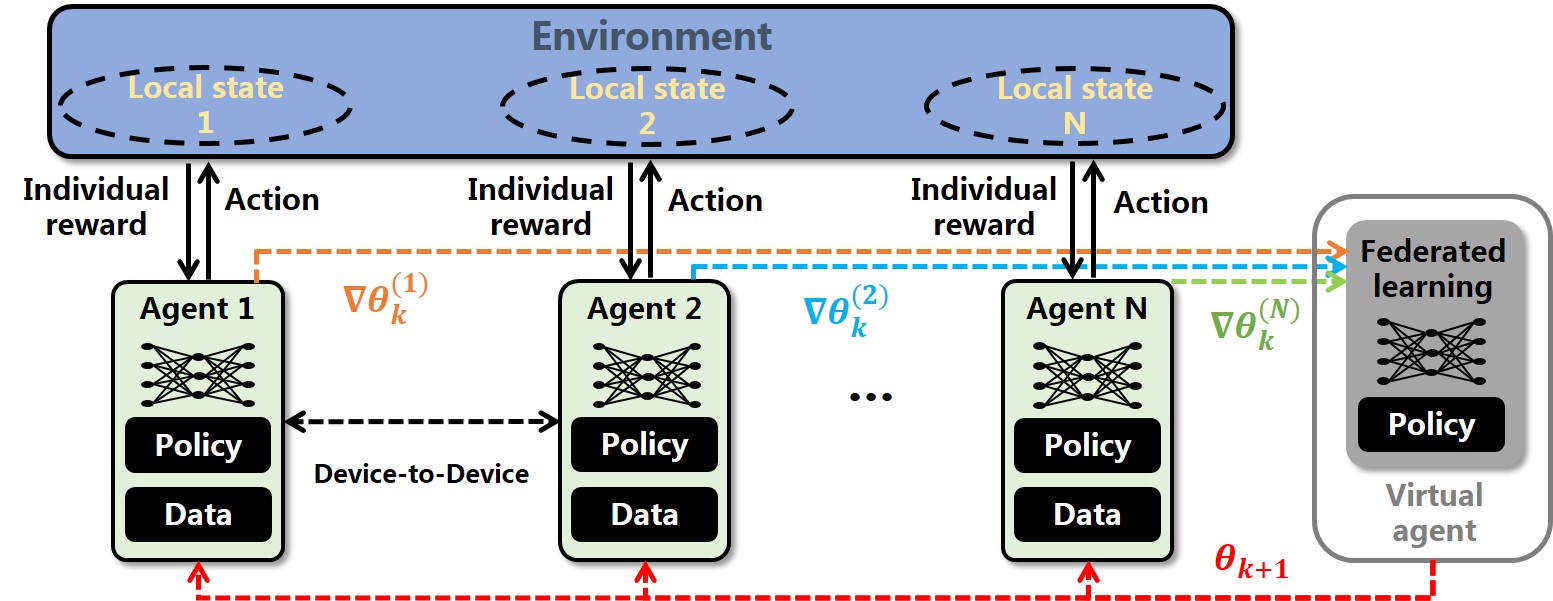}	
	\caption{Framework of FIRL.}
	\label{fig1}
	\vspace{-0.2cm}
\end{figure}

\section{System Model and Problem Formulation}
The framework of federated IRL (FIRL) is illustrated in Fig. \ref{fig1}. We assume that there are totally $N$ agents involved in this multi-agent distributed learning scenario. Each agent is designed to learn independently in the training phase, and operate autonomously in the execution phase via DRL. Similar to Dec-POMDP \cite{Deep2017Shayegan}, we suppose that each agent can only observe a local state of the global environment, and an individual reward will be returned to the agent immediately after an action being performed. Within this scenario, the role of a conventional central server is played by a virtual agent, which can be deployed at a remote cloud center or some high-powered local agent.  Meanwhile, to facilitate the multi-agent collaboration, we allow agents to exchange their local gradients with nearby collaborators through the device-to-device communication whenever needed.

We assume that each agent maintains a DRL model with parameterized policy $\pi(s, a;\theta)$, where $s \in \mathcal{S}$ is sampled from the local state space, $a \in \mathcal{A}$ is selected from the individual action space, $\pi : \mathcal{S} \times \mathcal{A} \to [0,1]$ is a stochastic policy, and $\mathbf{\theta} \in \mathbb{R}^{d}$, where $d$ indicates the dimension of parameters. According to stochastic policy gradient methods for the policy optimization process of DRL \cite{John2015Trust}, the model's parameters are updated by
\begin{equation}
	\small
	\mathbf{\theta}_{k+1} = \mathbf{\theta}_{k} - \eta\nabla_{\mathbf{\theta}_{k}}\mathbf{\mathcal{L}},
\end{equation}
where $k$ represents the index of policy iteration, $\eta$ is the learning rate with a reasonable step size, and $\mathbf{\mathcal{L}}$ represents the loss function that needs to be minimized. In this paper, on-policy policy gradient methods are used to improve agents' policy performance by optimizing the loss function $\mathbf{\mathcal{L}}$ which can be defined by $\mathbf{\mathcal{L}}(\theta) := -\log \pi(s,a;\theta) \Phi_{\pi}(s,a;\theta)$, and $\Phi_{\pi}(s,a;\theta)$ can be different expressions, such as the advantage function $A_{\pi}(s,a;\theta)$ \cite{712192} used in this paper. In addition, according to the well-known experience replay mechanism in DRL \cite{Volodymyr2015Human}, we define the sample used in the policy iteration as $\phi_{t} :=\ <s_t,\ a_t,\ r_t,\ s_{t+1}>$, where $t$ represents the time-stamp within an epoch, and $r_t$ is calculated by a reward function $\mathcal{R} : \mathcal{S} \to \mathbb{R}$. In practical applications, a certain number of samples are picked out at each iteration to comprise a mini-batch \cite{Volodymyr2015Human}. Therefore, the practical gradients used for training are written as
\begin{equation}
	\small
	g(\theta_k;\xi_k) = \frac{1}{|\xi_k|} \sum_{\phi_{t} \in \xi_k} \nabla \mathbf{\mathcal{L}}(\theta_k;\phi_{t}),
\end{equation}
where $\xi_k$ denotes the mini-batch at iteration $k$, and $|\xi_k|$ indicates its corresponding size. Combining the periodic averaging method in FL, we can obtain
\begin{equation}
	\small
	\mathbf{\theta}_{k+1}^{(i)} = \left\{
	\begin{array}{lcl}
		{\frac{1}{m} \sum_{i=1}^{m} \left[\mathbf{\theta}_{k}^{(i)} - \eta g(\theta_k^{(i)};\xi_k^{(i)})\right]}, &\text{$k\ \mathrm{mod}\ \tau = 0$}; \vspace{2ex}\\
		{\mathbf{\theta}_{k}^{(i)} - \eta g(\theta_k^{(i)};\xi_k^{(i)})}, &\text{otherwise},   
	\end{array}  
	\right. 
\end{equation}
where $\xi_k^{(i)}$ is the mini-batch from agent $i$, and $\tau$ indicates a predefined number of local updates in a period, which is specially designed for the agent with the fast training speed. The reason for this design is that this fast agent can collect the most samples in the learning environment, and thus has the greatest weight on the training model. To reduce the total training time, we also assume that only the agents which have finished at least one local update at the end of a period need to transmit their local gradients to the virtual agent, and set the maximal number of these involved agents as $m$, where $m \le N$. In the rest of this paper, we will focus on these $m$ agents and use the notation $g(\theta_k^{(i)})$ to represent $g(\theta_k^{(i)};\xi_k^{(i)})$. Besides, the maximal length of an epoch and the total number of epochs for training are denoted as $T$ and $U$, respectively. To facilitate the convergence speed of policy in DRL, an epoch is further uniformly divided into several steps, each of which generally contains a predefined number of sequential Markov state transitions. In particular, each step involves $P$ transitions at most as a mini-batch.

Considering the resources spent by agents in the training process of FIRL, we basically suppose that the main communication overheads required for an agent to transmit its model's local gradients to the virtual agent are $C_1$, while the main computation overheads to perform a local updating are $C_2$. Then, the system's resource cost can be formulated as
\begin{equation}
	\small
	\psi_0 = \sum_{i=1}^{m} \left(\frac{C_1TU}{\tau P} + \frac{C_2 \tau_i TU}{\tau P}\right),
	\label{psi1}
\end{equation}
where $\tau_i \in \{1,2,3,...,\tau\}$, for $i = 1,2,3,...,m$. On the other hand, according to stochastic policy gradient methods, for any epoch, the optimization objective of policy iteration is expressed as
\begin{equation}
	\small
	\pi_{\theta} = \mathop{\arg\min}_{\theta \in \mathbb{R}^d} \mathbb{E}_t\left[\mathbf{\mathcal{L}}(\pi_{\theta};\phi_{t})\right].
	\label{obj}
\end{equation}
According to $(1)$ and the definition of $\mathbf{\mathcal{L}}$, we accomplish the optimization objective in (\ref{obj}) by performing stochastic gradient descent (SGD) for the policy parameters $\theta$. Next, we will replace the notation $\pi_{\theta}$ with $\theta$ in the following discussions. Furthermore, we can define an empirical risk function as
\begin{equation}
	\small
	F(\theta) := \mathbf{\mathcal{L}}(\theta;\xi).
\end{equation}
In practical applications, since the empirical risk function $F(\theta)$ may be non-convex, the learning model's convergence may fall into a local minimum or saddle point. In our framework, similar to \cite{Wang2018Cooperative,Bottou2018OptimizationMF}, the expected gradient norm is used as an indicator of the model's convergence performance to guarantee that it falls into a stationary point
\begin{equation}
	\small
	\mathbb{E}\left[\frac{1}{K} \sum_{k=0}^{K-1} \left\|\nabla F(\bar{\theta}_k)\right\|^2\right] \le \psi_1,
\end{equation}
where $K$ represents the expected total number of policy iterations, and $K = UT/P$. $\psi_1$ denotes the error convergence bound and helps define a bounded value for a targeted optimal solution. $\|\cdot\|$ denotes the $\ell_2$ vector norm. $\bar{\theta}_k$ represents the model's average parameters at iteration $k$, which is also regarded as the final result at the end of each iteration. The model with $\bar{\theta}_k$ is maintained by the virtual agent and updated periodically. Accordingly, we define $\bar{\theta}_0$ as the model's initial parameters for all agents and can obtain
\begin{equation}
	\small
	\psi_2 := \mathbb{E}\left[\left\|\nabla F(\bar{\theta}_0)\right\|^2\right],
\end{equation}
where $\psi_2$ denotes the expected gradient norm of the initial model, which represents the initial convergence error and is gradually decreased by SGD. Finally, we can present the optimization objective of FIRL as
\begin{equation}
	\small
	\rm{maximize}\ \frac{\alpha(\psi_2 - \psi_1)}{\psi_0},
	\label{maxi}
\end{equation}
where $\alpha$ is a positive constant to reflect the relative importance of convergence performance and resource cost. Hereinafter, we will use (\ref{maxi}) as the targeted system's utility function, for which the optimization progress can help achieve a good balance between improving the model's convergence performance and reducing the required resource cost.

\section{Consensus in Periodic Averaging Method}
In this paper, the model's error convergence bound is inferred under the following general assumptions, which are similar to those presented in previous studies for distributed SGD \cite{Wang2018Cooperative,Bottou2018OptimizationMF}.

\vspace{0.5em}
\noindent
\rm\textbf{Assumption 1 (A1).}

\begin{itemize}
	\small
	\setstretch{1.4}
	\item[1.] (Smoothness): $\left\|\nabla F(\theta) - \nabla F(\theta^{'})\right\| \le L\left\|\theta - \theta^{'}\right\|$;
	\item[2.] (Lower bounded): $F(\theta) \ge F_{\mathrm{inf}}$;
	\item[3.] (Unbiased gradients): $\mathbb{E}_{\xi|\theta}\left[g(\theta)\right] = \nabla F(\theta)$;
	\item[4.] (Bounded variance): $\mathbb{E}_{\xi|\theta}\left[\left\|g(\theta) - \nabla F(\theta)\right\|^{2}\right] \le \beta\left\|\nabla F(\theta)\right\|^{2} + \sigma^{2}$,
\end{itemize}
where $L$ represents the Lipschitz constant, which implies that the empirical risk function $F$ is $L$-smooth \cite{Zhou2018Duality}. $F_{\mathrm{inf}}$ denotes the lower bound of $F$, and we suppose that it can be reached when $K$ is large enough. In addition, $\beta$ and $\sigma^2$ are both non-negative constants and are inversely proportional to the size of mini-batch \cite{Wang2018Cooperative}. Condition 3 and 4 in A1 on the bias and variance of the mini-batch gradients are customary for SGD methods \cite{Rui2020CPFed}. Specifically, the variance of the mini-batch gradients is bounded by Condition 4 through the value that fluctuates with the exact gradients rather than through a fixed constant as in previous studies \cite{Hao2018Parallel,Rui2020CPFed}, which thus sets up a looser restriction. 

During the training phase of DRL, it is common to keep the learning rate $\eta$ as a proper constant and decay it only when the performance saturates. In the following discussions, we will investigate the model's error convergence bound under a fixed learning rate. Since the original objective of the consensus algorithm is to make all the distributed nodes in an ad-hoc network reach a consensus, it can be used to reduce the variance of the mini-batch gradients from a cluster of agents. Therefore, we employ the consensus algorithm \cite{R2007Consensus} to improve the local updating process of each agent. Next, we will use the notation $g(\theta_{k}^{(i)}, e)$ to represent $g(\theta_{k}^{(i)})$, where $e$ denotes the index of local interaction, and $g(\theta_{k}^{(i)}, 0) = g(\theta_{k}^{(i)})$. To enable all the participating agents to reach a consensus successfully, we make the following assumption for the network of participating agents.

\vspace{0.5em}
\noindent
\rm\textbf{Assumption 2 (A2).} \textit{The network of participating agents with topology $G$ is a strongly connected undirected graph.}\\
Then, according to the consensus algorithm \cite{R2007Consensus}, we can obtain the following local interaction process of each agent
\begin{equation}
	\small
	g(\theta_{k}^{(i)}, e + 1) = g(\theta_{k}^{(i)}, e) + \epsilon \sum_{l \in \Omega_{i}}\left[g(\theta_{k}^{(l)}, e) - g(\theta_{k}^{(i)}, e)\right],
\end{equation}
where $\Omega_{i}$ represents the set of neighboring agents that are directly connected with agent $i$ in $G$. $\epsilon$ denotes the local interaction step size, which plays a similar role as the learning rate $\eta$ in the local interaction process. And $0 < \epsilon < 1/\Delta$, where $\Delta$ denotes the maximal degree of the graph and is defined by $\Delta := \max_i | \Omega_{i}| + 1$. Furthermore, we can present the update rule of the model's parameters under the consensus-based method by
\begin{equation}
	\small
	\theta_{k}^{(i)} = \bar{\theta}_{t_0} - \eta \sum_{s = t_0}^{k - 1} g(\theta_{s}^{(i)}, e);
	\label{conagent}
\end{equation}
\begin{equation}
	\small
	\bar{\theta}_{k} = \bar{\theta}_{t_0} - \eta \frac{1}{m} \sum_{i = 1}^{m} \sum_{s = t_0}^{k - 1} g(\theta_{s}^{(i)}, e),
	\label{convirtual}
\end{equation}
where $t_0 = z\tau$, $z \in \mathbb{N}$, and $t_0$ represents the index of the iteration, at which the virtual agent performs the latest periodic averaging before iteration $k$.
It can be observed that agents need to exchange their local gradients with nearby collaborators before performing the local updating, and these exchanged gradients are also used to update the model's average parameters. Note that (\ref{conagent}) and (\ref{convirtual}) also consider the delay of agents in obtaining local gradients, where $g(\theta_{k}^{(i)}, 0)$ equals to $\mathbf{0}$ for some agents at the beginning of local interaction. Thus, although local interactions occur synchronously between neighboring agents, it is not necessary for an agent to wait for all the neighbors to complete the calculation of local gradients, so as to reduce the training time. Finally, we can obtain the following theorem

\vspace{0.5em}
\noindent
\rm\textbf{Theorem 1 (T1).} \textit{Suppose the number of local updates for agent $i$ is $\tau_i$, and the total number of iterations $K$ is large enough, which can be divided by $\tau$. Under A1 and A2, if the learning rate $\eta$ satisfies} 
\begin{equation}
	\small
	\eta L (\frac{\beta}{m} + 1) - 1 + 2 \eta^{2} L^{2} \tau \beta + \eta^{2} L^{2} \tau (\tau + 1) \le 0,
	\label{t1}
\end{equation}
\textit{then the expected gradient norm after $K$ iterations is bounded by}
\begin{equation}
	\small
	\begin{aligned}
		\mathbb{E}\left[\frac{1}{K} \sum_{k=0}^{K-1} \left\|\nabla F(\bar{\theta}_k)\right\|^{2}\right]& \le \frac{2[F(\bar{\theta}_0) - F_{\mathrm{inf}}]}{\eta K}  +  \frac{\eta L \sigma^{2}}{m} \\ & +  \eta^{2} \sigma^{2} L^{2}(\tau + 1)\left[1 - \epsilon \mu_{2}(\mathbf{La})\right]^{2E},
		\label{lb5}
	\end{aligned}
\end{equation}
where $E$ represents the total number of local interactions before each local updating. $\mathbf{La}$ denotes the Laplace matrix of the graph $G$. $\mu_{2}(\mathbf{La})$ denotes the second smallest eigenvalue of $\mathbf{La}$, which is also called \textit{algebraic connectivity} \cite{R2007Consensus}. 

\noindent
\emph{Remarks:} The corresponding proofs are presented in the Appendix. Note that we leave the Appendix in the Supplemental Material due to the space limitation\footnote{The Supplemental Material is available at https://www.rongpeng.info/files/sup\_icc2022.pdf.}. It can be observed that (\ref{t1}) indicates an upper bound for the value of learning rate $\eta$. Moreover, the right side of (\ref{lb5}) shows that the model's error convergence bound (i.e.,  $\psi_1$) is determined by several key parameters. In particular, the convergence bound will decrease as the total number of iterations $K$ increases, and it can be also declined by the model's better initial parameters $\bar{\theta}_0$. Taking into account the multi-agent parallel training paradigm, increasing the number of participating agents $m$ in the framework of FIRL can  reduce the convergence bound, but this comes at the expense of more resource cost according to (\ref{psi1}) and may lead to a reduction in the system's utility value according to (\ref{maxi}). Besides, it can be observed that although the implementation of the periodic averaging method can reduce the communication overheads by many times, it will enlarge the convergence bound and may be harmful to the system's utility value. The part in square brackets of the third term on the right side of (\ref{lb5}) describes the topological properties of the graph comprised by agents and their connections through the algebraic connectivity and the local interaction step size, instead of an upper bound on the spectral radius of a stochastic matrix for interaction \cite{Multi-Stage2020Seyyedali}. In particular, since $0 < \mu_{2}(\mathbf{La}) \le \Delta$ and the equality holds only when $G$ is a fully connected graph, we can find that $0 < 1 - \epsilon \mu_{2}(\mathbf{La}) < 1$. Therefore, it can be inferred that the implementation of local interactions can further reduce the convergence bound greatly. In addition, a larger step size $\epsilon$ or a more densely connected network of participating agents can also help reduce this bound. In practical applications, a small number of local interactions can make the model's convergence bound decrease dramatically, such as $E = 2$.

On the other hand, considering the resource cost under the consensus-based method, we assume that the communication overheads required for an agent to exchange the mini-batch gradients with one of its neighbors are $W_1$, and the computation overheads required to perform a local interaction are $W_2$. Then the system's resource cost can be updated from (4) as
\begin{equation}
	\small
	\psi_3 = \sum_{i=1}^{m} \left[\frac{C_1TU}{\tau P} + \frac{C_2 \tau_i TU}{\tau P} + |\Omega_{i}|(W_1 +W_2)\frac{ETU}{P}\right],
\end{equation}
where $|\Omega_{i}|$ represents the size of $\Omega_{i}$. Here the possible collision or interference issues in actual communication are omitted. Compared with $\psi_0$ defined in (\ref{psi1}), we can observe that the resource cost under the consensus-based method is increased, resulting from the additionally embedded local interactions. However, since the model's convergence bound is reduced at the same time, the system's utility value may be improved in certain cases. For example, suppose that the overheads required for participating agents to implement the device-to-device communication are much less than those to transmit the same messages to the remote virtual agent, then the effectiveness of the  consensus-based method will be clearly demonstrated. In addition, the network of participating agents can be also set up as a sparse graph to decrease the number of local interactions between agents, so as to further reduce the total resource cost.

\begin{table*}
	\footnotesize
	\centering
	\caption{Numerical simulation results.}
	\label{tb2}
	\begin{tabular}{c|c|c|c|c}
		\toprule[1.1pt]
		Methods & Local Updates & Communication overheads& Computation overheads& Expected gradient norm\\
		\hline
		FIRL & $\tau = 1$     & 21000 $C_1$ & 21000 $C_2$   & 1.5590 \\
		FIRL & $\tau = 10$    & 2100 $C_1$  & 21000 $C_2$   & 6.3421  \\
		FIRL & $\tau = 15$    & 1400 $C_1$  & 21000 $C_2$   & 9.6069  \\
		FIRL & $\tau = 10\thicksim 15$ & 1400 $C_1$  & 19000 $C_2$   & 10.1892 \\
		FIRL & $\tau = 5\thicksim15$    & 1400 $C_1$  & 16200 $C_2$   & 8.7182  \\
		FIRL & $\tau = 1\thicksim15$  & 1400 $C_1$  & 12600 $C_2$   & 7.6476  \\
		FIRL\_C & $\tau = 10$, $e=1$, $\mu_{2}=1.4384$  & 2100 $C_1$ + 78000 $W_1$  & 21000 $C_2$ + 78000 $W_2$  & 3.9577  \\
		FIRL\_C & $\tau = 10$, $e=1$, $\mu_{2}=2.5188$  & 2100 $C_1$ + 96000 $W_1$ & 21000 $C_2$ + 96000 $W_2$  & 1.5560  \\
		FIRL\_C & $\tau = 10$, $e=2$, $\mu_{2}=1.4384$  & 2100 $C_1$ + 156000 $W_1$ & 21000 $C_2$ + 156000 $W_2$  & 2.6094  \\
		FIRL\_C & $\tau = 1\thicksim10$, $e=1$, $\mu_{2}=1.4384$  & 2100 $C_1$ + 78000 $W_1$ & 15600 $C_2$ + 78000 $W_2$ & 3.4070  \\
		\bottomrule[1.1pt]
	\end{tabular}
\end{table*}

\section{Simulation Results}
\begin{figure}
	\centering
	\includegraphics[width=0.17\textwidth]{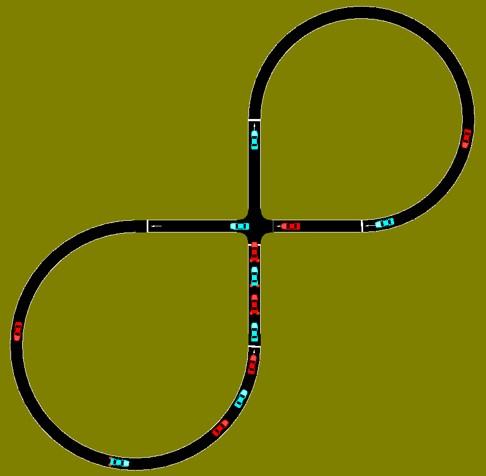}	
	\caption{Figure Eight.}
	\label{fig3}
\end{figure}

In this part, we talk about the simulation results after applying the consensus-based method. The simulation scenario is taken from \cite{VinitskyBenchmarks}, which is released as a new benchmark in traffic control through the implementation of DRL to create controllers for mixed-autonomy traffic. As illustrated in Fig. \ref{fig3}, the ``Figure Eight" simulation scenario in this benchmark is selected in this paper to verify the effectiveness and efficiency of the developed method. Specifically, there are totally 14 vehicles running circularly along a one-way lane that resembles the shape of a figure ``8". An intersection is located at the lane, and each vehicle must adjust its acceleration to pass through this intersection in order to increase the average speed of the whole vehicle team. Note that slamming on the brakes will be forced on the vehicles that are about to crash, and the current epoch will be terminated once the collision occurs. Furthermore, the ``Figure Eight" scenario is modified in this paper to assign the related local state of the global environment to each vehicle, including the position and speed of its own, the vehicle ahead and behind. Depending on its local state, each vehicle needs to optimize its acceleration, which is a normalized continuous variable between $-1$ and $1$. All the involved vehicles are the same, and unless additional controllers are assigned, these vehicles are controlled by the underlying simulation of urban mobility (SUMO) in the same mode, which is an open source with highly portable and widely used traffic simulation package \cite{VinitskyBenchmarks}. In the ``Figure Eight" scenario, half of the vehicles are assigned the DRL-based controllers. We take the normalized average speed of all vehicles at each iteration as the individual reward, which is assigned to each training vehicle after its action being performed. And unless otherwise specified, the DRL-based controllers are optimized through the proximal policy optimization (PPO) algorithm \cite{John2017Proximal}. 

\begin{figure}
	\centering
	\includegraphics[width=0.4\textwidth]{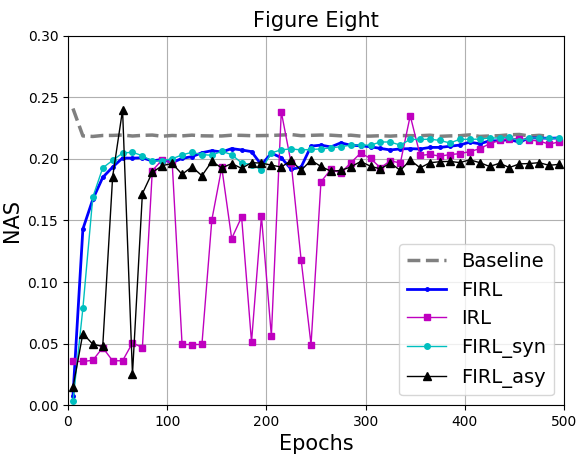}	
	\caption{Effectiveness of FL in IRL.}
	\label{figa}
	\vspace{-0.2cm}
\end{figure}

In Fig. \ref{figa}, we present the simulation performance of naive IRL and FIRL (IRL with naive FL). Here, NAS denotes the normalized average speed of all vehicles during an entire testing epoch. The test is performed every 10 epochs and takes the average of five repetitions. We provide the performance of all vehicles controlled by the underlying SUMO as the optimal baseline. In Fig. \ref{figa}, we show that FL can clearly improve the policy performance of IRL in terms of training efficiency and stability, while verify the necessity of this combination. We also test the effect of transmission delay in FL on the performance of IRL in the method FIRL\_syn and FIRL\_asy. Both methods introduce the delay of 2 epochs (i.e. 300 seconds) to the up-link and down-link between each agent and the virtual agent. However, agents in FIRL\_syn can update their parameters every iteration, while agents in FIRL\_asy can only update their parameters every 2 epochs, which means a worser communication. As indicated in Fig. \ref{figa}, a smaller frequency of agents updating their parameters may decline the effectiveness of FL in IRL.

In Table \ref{tb2}, we present a summarized comparison on the numerical simulation results to verify the effectiveness and efficiency of the developed method. In particular, we set $T=1500$, $U=500$, $P=250$, and $\eta = 1e-4$. The expected gradient norm is calculated upon a predetermined sample set, which is comprised by samples that are uniformly collected during the learning model's training process when $\tau = 1$. In addition, the expected gradient norm is calculated whenever the model's average parameters (i.e., $\bar{\theta}_{k}$) are updated, and its final value is averaged across the entire training process of the corresponding learning model. In Table \ref{tb2}, we present the performance of FIRL with different local updates in a period. The notation ``$\tau=10\sim15$" denotes that the numbers of local updates from agents are uniformly distributed between $10$ and $15$. We can observe a trend that the learning model's error convergence bound will increase as the number of local updates $\tau$ increases, which is consistent with our previous discussions. In Table \ref{tb2}, we also present the performance of FIRL under the consensus-based method (i.e. FIRL\_C). On the premise of strong connectivity, the topology of agents' network with $\mu_{2}=1.4384$ is constructed by $3 \sim 4$ random connections from each agent to nearby collaborators, while these connections are increased to $4 \sim 6$ when $\mu_{2}=2.5188$. Note that there is only one connection between any pair of agents. We can observe that being consistent with the conclusions in T1, the error convergence bound of targeted model is decreased when the local interactions are considered, even when the delay in local updating is taken into account (i.e., in case $\tau = 1\sim10$). In addition, the topology with either a larger $\mu_{2}$ (i.e., more connections) or more local interactions (i.e., greater $e$) generally obtains better training performance. Compared to FIRL with $\tau = 10$, FIRL\_C with $\tau = 10$ requires the same amount of resource cost in FL, and although there are more resources required in local interactions, its expected gradient norm is lower. Thus, the system utility value can be also declined when $W_1$ and $W_2$ are small in FIRL\_C. 

\section{Conclusions}
This paper has taken advantage of the paradigm of FL to improve the policy performance of IRL agents. Meanwhile, considering the excessive communication overheads generated between agents and a central server in FL, this paper builds the framework of FIRL on the basis of the periodic averaging method. Moreover, to reach a good balance between reducing the system's communication and computation overheads and improving the model's convergence performance, a novel utility function has been proposed. Furthermore, to improve the system's utility value, we have put forward a consensus-based optimization scheme on top of the periodic averaging method. By analyzing the theoretic convergence bounds and performing extensive simulations, both effectiveness and efficiency of the developed method have been verified. 

For future works, we plan to implement the proposed optimization method in the real-world application scenarios. In practice, when there is a large number of participating agents in the system, multiple virtual central agents may exist simultaneously and their organization tends to be hierarchical. This is a more complex scenario and requires more careful considerations on the optimization methods. Besides, whether the common shared deep neural network can be applied to heterogeneous tasks faced by different agents remains an open question that is worth exploring. 

\bibliographystyle{IEEEtran}
\bibliography{ref}
\end{document}